\documentclass[journal]{IEEEtran}

\usepackage{latexsym,,amssymb,amsmath,graphicx,epsf,cite,bbm,float}
\usepackage{ifpdf}
\usepackage{epstopdf}

\usepackage{algorithm,algorithmic,hyperref}
\usepackage{amsmath,amssymb,bm}
\usepackage{amsfonts,dsfont,color,bbm,subcaption}

\newcommand{\be}{\begin{equation}}
\newcommand{\ee}{\end{equation}}
\newcommand{\bea}{\begin{eqnarray}}
\newcommand{\eea}{\end{eqnarray}}

\newcommand{\MB}{\left[\begin{array}}
\newcommand{\ME}{\end{array}\right]}

\newcommand{\ei}{\end{itemize}}
\newcommand{\bi}{\begin{itemize}}

\usepackage{amsmath,amsthm}

\newcommand{\E}{\mathbb{E}}
\newcommand\Tau{\mathcal{T}}

\newtheorem{theorem}{Theorem}

\newtheorem{lemma}[]{Lemma}

\newtheorem{corollary}[]{Corollary}

\newtheorem{remark}[]{Remark}

\begin{document}

\title{Generalized Translation and Scale Invariant Online Algorithm for  Adversarial Multi-Armed Bandits} 
\author{Kaan Gokcesu, Hakan Gokcesu}
\maketitle

\flushbottom

\begin{abstract}
	We study the adversarial multi-armed bandit problem and create
	a completely online algorithmic framework that is invariant under arbitrary translations and scales of the arm losses. We study the expected performance of our algorithm against a generic competition class, which makes it applicable for a wide variety of problem scenarios. Our algorithm works from a universal prediction perspective and the performance measure used is the expected regret against arbitrary arm selection sequences, which is the difference between our losses and a competing loss sequence. The competition class can be designed to include fixed arm selections, switching bandits, contextual bandits, or any other competition of interest. 
	The sequences in the competition class are generally determined by the specific application at hand and should be designed accordingly. Our algorithm neither uses nor needs any preliminary information about the loss sequences and is completely online. Its performance bounds are the second order bounds in terms of sum of the squared losses, where any affine transform of the losses has no effect on the normalized regret. 
\end{abstract}

\section{Introduction}\label{sec:intro}
\subsection{Preliminaries}
In machine learning literature \cite{jordan2015machine,mohri2018foundations}, the area of online learning \cite{shalev2011online} is heavily investigated in various fields from game theory \cite{chang,tnnls1}, control theory \cite{sw2,sw4,tnnls3}, decision theory \cite{tnnls4,freund1997} to computational learning theory \cite{comp1,comp2}. Because of the heavily utilized universal prediction perspective \cite{merhav}, it has been considerably applied in data and signal processing \cite{sw3,signal2,moon,signal1,sw5,gokcesu2018adaptive}, especially in sequential prediction and estimation problems \cite{gHierarchical,singer,ozkan,singer2} such as the problem of density estimation and anomaly detection \cite{gAnomaly,willems,coding1,coding2,gDensity}. 
Some of its most prominent applications are in multi-agent systems \cite{sw1,vanli,tekin2014distributed} and specifically in reinforcement learning problems \cite{auerExp,bandit1,bandit2,audibert,tekin2,gBandit,exptrade,auer,auerSelf,reinInt,mannucci2017safe}.

In these types of applications, we encounter the fundamental dilemma of exploration-exploitation trade-off, which is most throughly studied in the multi-armed bandit problem \cite{cesa-bianchi}. To that end, study of the multi-armed bandit problem has received considerable attention over the years \cite{bandit2,cesa-bianchi,audibert,auer,auerExp,banditTNN,zheng,gBandit}, where the goal is to minimize or maximize some loss or reward, respectively, in a problem environment by sequentially selecting one of $M$ given actions \cite{cesabook}.

The multi-armed bandit is widely considered to be the limited feedback version of the well studied prediction with expert advice \cite{signal2,signal1,singer,moon,singer2}. Due to the nature of the problem, only the loss of the selected arm is observed (while others remain hidden). The multi-armed bandit problem has attracted significant attention due to its applicability in a wide range of problem settings from online advertisement \cite{li2010contextual} and recommender systems \cite{tekin2014distributed,tang2014ensemble,luo2015nonnegative,li2011unbiased} to clinical trials \cite{hardwick1991bandit} and cognitive radio \cite{lai2008medium,gai2010learning}.

\subsection{Adversarial Multi-armed Bandit Problem}
We study the multi-armed bandit problem in an online setting, where we sequentially operate on a stream of observations from an adversarial environment \cite{huang2011adversarial}, i.e., we have no statistical assumptions on the loss sequence. To this end, we investigate the multi-armed bandit problem from a competitive algorithm perspective \cite{merhav,vural2019minimax,vovk,littlestone1994,neyshabouri2018asymptotically,vovk1998,gokcesu2020recursive}. 

In the competitive perspective, the performance is defined with respect to a competition class of arm selection strategies. For any sequence of losses, the goal of an algorithm is to achieve a cumulative loss that is as close to the cumulative losses of the arm selection sequences in the competition (e.g., for fixed bandit arm competitions, we compare against the bandit arm with the best cumulative loss) \cite{cesa2007}. The difference between the cumulative loss of the algorithm and the best arm selections on the same loss sequence is called 'regret' \cite{cesabook}. 

In the competitive algorithm perspective, one does not need to explicitly know the actions corresponding to each bandit arm available. Each bandit arm can either be separately running black-box algorithms that learn throughout time instead of some specific action. The only prior knowledge is on the number of the bandit arms (whatever they may be), so that an algorithm knows which arm corresponds to which action. In the competitive framework, the action at any time $t$ is decided upon observed sequential performances. 

The adversarial multi-armed bandit problem where the competition is against the best fixed arm has a regret lower bound of $\Omega(\sqrt{MT})$ for $M$ bandit arms in a $T$ round game \cite{cesa-bianchi}. When competing against the arm sequences in an arbitrary competition class (as opposed to the best fixed arm), the regret lower bound implies a minimax bound of $\Omega(\sqrt{W_*MT})$, where $W_*$ is the complexity of the competition arm sequence. This complexity can be dependent on either the number of switches in the sequence \cite{cesa-bianchi,auer,auerExp,audibert,doubling_trick}, the number of contextual regions \cite{willems1995context,sadakane2000implementing,willems1996context,csiszar2006context,dumont2014context,kozat2007universal,vanli2014comprehensive,ozkan2016online}, or any other complexity definition that implies a prior on the arm sequences \cite{comp2,gokcesu2020generalized}. 
With some alterations, these state of the art algorithms achieve an expected minimax regret upper bound $O(\sqrt{W_*MT})$ when $W_*$ is known a priori. They are also able to attain an expected regret upper bound of $O(W_*\sqrt{MT})$ when the complexity $W_*$ of the best arm sequence is not known a priori. They can also be utilized to achieve a regret bound of $O(\sqrt{WMT})$ when competing against the arm sequence with complexities upper bounded by $W$.
However, they all lack in the aspect that they simply assume the losses are bounded (generally in the $[0,1]$ region). Because of the lack of adaptivity in their algorithms, their regret bounds may be in the order of $O(L^2\sqrt{WMT})$, where $L$ is an unknown upper bound on the losses.

\subsection{Translation and Scale Invariant Regret Bounds}
The search for fundamental regret bounds, i.e., translation and scale invariance, has been popular in the literature for a few decades, especially for the problem of prediction with expert advice. Against fixed competition, the most straightforward approach via the exponentially weighted average forecaster of \cite{littlestone1994,vovk1998} provides a zeroth order regret bound (the regret is dependent on the universal loss bound and the number of rounds). In one-sided games, when all losses have the same sign, \cite{freund1997} showed that the algorithm of \cite{littlestone1994} can obtain a first order regret bound (where the regret is dependent on the sum of the losses). A direct analysis on the signed games in \cite{allenberg2004} uncovers that weighted majority actually achieves the first order regret without a need for one-sidedness in the losses. These approaches are all scale invariant, however, they do not have translation or parameter-freeness (since some information about the losses are needed a priori). 

The work in \cite{cesa2007} solves these shortcomings by creating second order regret bounds (where the regret is dependent on the sum of squared losses) for signed games (thus improving upon the previous existing bounds), and simultaneously eliminating the need for any a priori knowledge. Their algorithm is translation, scale invariant and also parameter-free. However, their competition class is limited and mainly focused on fixed competition. Although, there exist variants in literature to deal with different applications, competing against arbitrary bandit arm selection sequences is nontrivial (because of the problem's nature) unless you treat each sequence as a bandit arm itself, which would be difficult to implement for a scenario with a large competition class.

The approach in \cite{gokcesu2020generalized} addresses this issue by extending the second order regret bounds to a generalized framework that is able to arbitrarily compete against different choices of competitions in the problem of prediction with expert advice. However, their results are not applicable to the bandit setting because of the limited feedback.
\subsection{Contributions and Organization}
To this end, we improve upon the previous works to provide an algorithmic framework to compete against arbitrary bandit arm selection sequences with translation and scale invariant regret bounds. Much like \cite{gokcesu2020generalized}, our algorithmic framework can straightforwardly implement the desired competition class in a scalable and tractable manner. Since, in the competitive algorithm perspective, we do not need to explicitly know the actions (bandit arms) we are presented with, the only prior knowledge we need about the bandit arms is that there are $M$ options that we can select from, and what kind of bandit arm selection sequences we want to compete against (the competition class). Our algorithm sequentially selects the arm solely based on the past performances and is parameter-free. 

The organization of the paper is as follows. In Section \ref{sec:problem}, we first describe the bandit arm selection problem. Then, in Section \ref{sec:method}, we detail the methodology and our algorithmic framework. We provide the performance results and regret analysis in Section \ref{sec:regret}. We conclude with some important remarks in Section \ref{sec:conc}.

\section{Problem Description}\label{sec:problem}
In this paper, we study the adversarial multi-armed bandit problem where we have $M$ bandit arms such that $m\in\{1,\ldots,M\}$ and randomly select one of them at each round $t$. We select our bandit arms according to our selection probabilities
\begin{align}
	q_t\triangleq[q_{t,1},\ldots,q_{t,M}],\label{eq:qt}
\end{align}
and our selection is $i_t\in\{1,\ldots,M\}$ randomly drawn from it such that
\begin{align}
	i_t\sim q_t.
\end{align}
Based on our online selection 
\begin{align}
	\{i_t\}_{t\geq1},\enspace i_t\in\{1,2,\ldots,M\},
\end{align}
we incur the loss of the selected arm
\begin{align}
	 \{{l_{t,i_t}}\}_{t\geq1},
\end{align}
where we do not assume anything about the losses before selecting our bandit arm at time $t$. Because of the bandit setting, we do not observe the other losses $\{{l_{t,{m\neq i_t}}}\}_{t\geq1}$ (i.e., they are hidden).

In a $T$ round game, we define $I_T$ as the row vector containing the user selections up to time $T$ as 
\begin{align}
	I_T=[i_1,\ldots,i_T],
\end{align}
and the loss sequence of $I_T$ as
\begin{align}
	L_{I_T}=[l_{1,{i_1}},\ldots,l_{T,i_T}].
\end{align}
Similarly, we define $S_T$ as the row vector representing a deterministic bandit arm selection sequence of length $T$ as
\begin{align}
	S_T=[s_1,\ldots,s_T].\label{eq:St}
\end{align}
such that each $s_t\in\{1,2,\ldots,M\}$ for all $t$. In the rest of the paper, we refer to each such deterministic bandit arm selection sequence, $S_T$, as a competition. Hence, the loss sequence of the competition $S_T$ is
\begin{align}
	L_{S_T}=[l_{1,{s_1}},\ldots,l_{T,s_T}].
\end{align}
We denote the cumulative loss at time $T$ of $I_T$ by 
\begin{align}
	{C_{I_T}=sum(L_{I_T})=\sum_{t=1}^T l_{t,i_t}},
\end{align}
and similarly the cumulative loss at time $T$ of $S_T$ by 
\begin{align}
{C_{S_T}=sum(L_{S_T})=\sum_{t=1}^T l_{t,s_t}}.
\end{align}
Since we assume no statistical assumptions on the loss sequence, we define our performance with respect to a competition $S_T$ that we want to compete against.
We use the notion of regret to define our performance against any competition $S_T$ as
\begin{align}
R_{S_T}&\triangleq C_{I_T}-C_{S_T}=\sum_{t=1}^T l_{t,i_t}-\sum_{t=1}^T l_{t,s_t},\label{RT1}
\end{align}
where we denote the regret accumulated in $T$ rounds against $S_T$ as $R_{S_T}$. Our goal is to create an algorithm with expected regret bounds against $S_T$ that depends on how hard it is to learn the competition $S_T$.

\section{The Algorithm}\label{sec:method}

	\begin{algorithm}[!t]
		\caption{Generalized Algorithm for Bandit Arm Selection}\label{alg:framework}
		\small{\begin{algorithmic}[1]
				\FOR{$t=1$ \TO $T$}
				\FOR{$m\in\{1,\ldots,M\}$}
				\STATE $$q_{t,m}=(1-\epsilon_{t})p_{t,m}+\epsilon_{t}\frac{1}{M}$$
				\ENDFOR
				\STATE Select $i_t\in\{1,\ldots,M\}$ with $q_t=[q_{t,1},\ldots,q_{t,M}]$
				\STATE Receive $\phi_{t}=[\phi_{t,1},\ldots,\phi_{t,M}]$
				\FOR{$\lambda_t\in\Omega_t$}
				\STATE $$z_{\lambda_t}=w_{\lambda_t}\exp(-\eta_{t-1}\phi_{t,\lambda_t(1)})$$
				\ENDFOR
				\FOR{$\lambda_{t+1}\in\Omega_{t+1}$}
				\STATE $$w_{\lambda_{t+1}}=\sum_{\lambda_{t}\in\Omega_t}\Tau(\lambda_{t+1}|\lambda_t)z_{\lambda_t}^{\frac{\eta_{t}}{\eta_{t-1}}}$$
				\ENDFOR
				\FOR{$m\in\{1,\ldots,M\}$}
				\STATE $$w_{t+1,m}={\sum_{\lambda_{t+1}(1)=m}w_{\lambda_{t+1}}}$$
				\ENDFOR
				\FOR{$m\in\{1,\ldots,M\}$}
				\STATE $$p_{t+1,m}=\frac{w_{t+1,m}}{\sum_{m'=1}^{M}w_{t+1,m'}}$$
				\ENDFOR
				\ENDFOR
		\end{algorithmic}}
	\end{algorithm}
	
	The design of our algorithm starts similarly with \cite{gokcesu2020generalized}, where we use equivalence classes \cite{gBandit} to efficiently combine the arm selection sequences $S_t$ at time $t$.
	
	\subsection{Universal Combination to Create Arm Probabilities}
	Our algorithm works by implicitly assigning a weight $w_{S_t}$ to each of the bandit arm sequences $S_t$. Using these weights, we create the arm weights $w_{t,m}$. We find the sequences among all the sequences that suggest arm $m$ at time $t$ and sum their weights to create the weight of the arm $m$ at time $t$, i.e.,
	\begin{align}
	w_{t,m}\triangleq\sum_{S_t(t:t)=m}^{} w_{S_t},\label{wmt}
	\end{align}
	where $S_t(i\!:\!j)$ is the vector consisting of $i^{th}$ through $j^{th}$ elements of $S_t$, e.g., $S_t(t:t)=s_t$, which is the bandit arm selection of the sequence $S_t$ at time $t$. The sequences are combined according to their weights $w_{S_t}$ to intrinsically achieve the performance of the best sequence because of its universal perspective \cite{merhav}. By normalization, we construct the algorithmic probabilities, i.e.,
	\begin{align}
	p_{t,m}=\frac{w_{t,m}}{\sum_{m'}w_{t,m'}},\label{pmt}
	\end{align}
	and selection probabilities $q_{t,m}$ are given by mixing $p_{t,m}$ with a uniform distribution (which was unnecessary in \cite{gokcesu2020generalized}) as 
	\begin{align}
		q_{t,m}=(1-\epsilon_t)p_{t,m}+\epsilon_t\frac{1}{M},\label{qtm}
	\end{align}
	where $\epsilon_t$ is a time dependent parameter.
	
	\subsection{Equivalence Classes for Efficient Calculation}
	We point out that the construction of $p_{m,t}$ in \eqref{pmt} directly depends on $w_{S_t}$ in lieu of \eqref{wmt}. These weights are implicitly calculated with an equivalence class approach, where we mutually update certain arm sequence weights \cite{gBandit,gokcesu2020generalized}.
	To create the equivalence classes, we first define a class parameter $\lambda_t$ as
	\begin{align}
		\lambda_t=[m, \ldots],\label{lamt}
	\end{align}
	where the first parameter $\lambda_t(1)$ is arbitrarily set as the bandit arm selection $m$ at time $t$. Together with the omitted parameters in \eqref{lamt}, $\lambda_t$ determines the sequences that are included in its equivalence class, i.e., the equivalence class with parameters $\lambda_t$ includes all sequences $S_t$ whose behavior match with the parameters $\lambda_t$ as a whole. The parameters of $\lambda_t$ determine how many equivalence classes we have, and how many arm sequences each class represents. We define $\Omega_t$ as the vector space including all possible $\lambda_t$ vectors as
	\begin{align}
		\lambda_t\in\Omega_t, \enspace\forall\lambda_t.\label{omegat}
	\end{align}
	We point out that $\Omega_t$ may not necessarily represent all possible sequences at time $t$, but instead the sequences of our interest, which we want to compete against. We also define $\Lambda_t$ as the parameter sequence up to time $t$ for an arbitrary sequence as
	\begin{align}
		\Lambda_t\triangleq\{\lambda_1,\ldots,\lambda_t\},\label{Lamt}
	\end{align} 
	where each sequence $S_t$ will correspond to only one $\Lambda_t$. 
	We define $w_{\lambda_t}$ as the weight of the equivalence class parameters $\lambda_t$ at time $t$. The weight of an equivalence class is simply the summation of the implicit weights of the sequences whose behavior conforms with its class parameters $\lambda_t$, such that
	\begin{align}
	w_{\lambda_t}=\sum_{F_\lambda(S_t)=\lambda_t}^{}w_{S_t},\label{wlt}
	\end{align}
	where $F_\lambda(\cdot)$ is the mapping from sequences $S_t$ to the auxiliary parameters $\lambda_t$, which transforms the definition in \eqref{wmt} to
	\begin{align}
	w_{t,m}=\sum_{\lambda_t(1)=m}^{} w_{\lambda_t}.\label{wmt2}
	\end{align}
	Similar to \cite{gokcesu2020generalized}, we update the weights $w_{\lambda_t}$ using the following two-step approach. At first, we define an intermediate variable $z_{\lambda_t}$ (which incorporates the exponential performance update as in the exponential weighting algorithm \cite{cesabook}) such that
	\begin{align}
	z_{\lambda_t}\triangleq w_{\lambda_t}e^{-\eta_{t-1}\phi_{t,\lambda_t(1)}},\label{zlt}
	\end{align}
	where $\phi_{t,m}$ is a measure of the bandit arms performance, which we discuss more in the next section.
	Secondly, we create a probability sharing network among the equivalence classes (which also represents and assigns a weight to every individual sequence $S_t$ implicitly) at time $t$ as
	\begin{align}
	w_{\lambda_{t+1}}=\sum_{\lambda_t\in\Omega_t}\Tau(\lambda_{t+1}|\lambda_t)z_{\lambda_t}^{\frac{\eta_t}{\eta_{t-1}}},\label{wlt+}
	\end{align}
	where $\Tau(\lambda_{t+1}|\lambda_t)$ is the transition weight from the class parameters $\lambda_t$ to $\lambda_{t+1}$ such that $\sum_{\lambda_{t+1}\in\Omega_{t+1}}\Tau(\lambda_{t+1}|\lambda_t)=1$ (which is a probability distribution itself). The power normalization on $z_{\lambda_{t}}$ is necessary for adaptive learning rates \cite{gokcesu2020generalized}. A summary of the method is given in \autoref{alg:framework}.
	
\section{Parameter Design and Regret Analysis}\label{sec:regret}
	In this section, we study the performance of our algorithm. We first provide a summary of some important notations and definitions, which will be heavily used. Then, we study the regret bounds by successively designing the learning rates $\eta_t$, the performance measures $\phi_{t,m}$ and the uniform mixture coefficients $\epsilon_{t}$.
	\subsection{Notations and Definitions}\label{secsec:not}
	\begin{enumerate}
		\item $q_{t,m}$ is the probability of selecting $m$ at $t$ as in \eqref{qtm}.
		\item $\E_{f_{t,m}}[x_{t,m}]$ is the convex sum of $x_{t,m}$ with the coefficients $f_{t,m}$, i.e., $\sum_{m=1}^{M}f_{t,m}x_{t,m}$.
		\item $\E_{f_{t}}[x]$ is the expectation of $x$ when $i_t$ is drawn from $f_t$.
		\item $\E[x]$ is the expectation of $x$ when $i_t$ is drawn from $q_{t,m}$.
		\item $\eta_t$ is the learning rate used in \eqref{zlt}.
		\item $\phi_{t,m}$ is the performance metric used in \eqref{zlt}.
		\item $d_t\triangleq \enspace\max_m\phi_{t,m}-\min_m\phi_{t,m}$.
		\item $v_t\triangleq \enspace\E_{p_{t,m}}\phi_{t,m}^2$.
		\item $D_t\triangleq\max_{1\leq t' \leq t}d_t,$.
		\item $V_t\triangleq\sum_{t'=1}^t v_t$.
		\item $e$ is Euler's number.
		\item $\log(\cdot)$ is the natural logarithm. 
		\item $\lambda_t$ is an equivalence class parameter at time $t$ as in \eqref{lamt}.
		\item $\Omega_t$ is the set of all $\lambda_t$ at time $t$ as in \eqref{omegat}.
		\item $\Lambda_T\triangleq\{\lambda_t\}_{t=1}^T$ as in \eqref{Lamt}.
		\item $z_{\lambda_{t}}$ is as in \eqref{zlt}.
		\item $\Tau(\cdot|\cdot)$ is the transition weight used in \eqref{wlt+}.
		\item $\Tau(\{\lambda_t\}_{t=1}^T)\triangleq\prod_{t=1}^T\Tau(\lambda_t|\lambda_{t-1})$.		
		\item $W(\Lambda_T)\triangleq \log(\max_{1\leq t\leq T}|\Omega_{t-1}|)-\log(\Tau(\Lambda_T))$, which corresponds to the complexity of a competition.
\end{enumerate}

	\subsection{Performance Analysis}
	The performance analysis starts the same as in \cite{gokcesu2020generalized} with the difference $\phi_{t,m}\geq 0, \forall t,m$. Hence, we have the following. 
	
	\begin{lemma}\label{thm:bound}
		When using \autoref{alg:framework}, we have
		\begin{align*}
			\sum_{t=1}^T\left(\E_{p_{t,m}}\phi_{t,m}-\phi_{t,\lambda_t(1)}\right)\leq& \frac{1}{2}\sum_{t=1}^T\eta_t\E_{p_{t,m}}\phi_{t,m}^2\\
			&+\sum_{t=1}^T\left(1-\frac{\eta_t}{\eta_{t-1}}\right)d_t\\
			&+\frac{\log(\max_{1\leq t\leq T}|\Omega_{t-1}|)}{\eta_{T-1}}\\
			&-\frac{1}{\eta_{T-1}}\log(\Tau(\Lambda_T)),
		\end{align*}
		where $\Tau(\Lambda_T)=\Tau(\{\lambda_t\}_{t=1}^T)$; $\phi_{t,m}\geq 0$, for all $t,m$; $\eta_t$ is non-increasing with $t$.
		\begin{proof}
			The proof follows from \cite{gokcesu2020generalized}, where we use the inequality $e^{-x}\leq 1-x+\frac{1}{2}x^2$ for $x\geq 0$ (from Taylor series).
		\end{proof}
	\end{lemma}
	\autoref{thm:bound} provides us an upper bound on the cumulative difference on the performance variable $\phi_{t,m}$ (possibly related to the regret) in terms of the learning rates $\eta_t$. The selection of the learning rates drastically affects the upper bound and should be chosen with care. 
	\subsection[Designing the Learning Rates]{Designing the Learning Rates $\eta_t$}
	Differently from \cite{gokcesu2020generalized}, we set the following learning rates
	\begin{align}
		\eta_t=\frac{\gamma}{\sqrt{V_t+D_t^2}},\label{etat}
	\end{align}
	which are non-increasing and $\gamma$ is a user-set parameter.
	
	\begin{lemma}\label{thm:ntphi2}
		When using \autoref{alg:framework}, $\eta_t$ as in \eqref{etat}, we have
		\begin{align*}
			\frac{1}{2}\sum_{t=1}^T\eta_t\E_{p_{t,m}}\phi_{t,m}^2\leq \gamma\sqrt{V_T},
		\end{align*}
	where $\gamma$ is a user-set parameter.
		\begin{proof}
			From the definitions of $v_t$, $V_t$ and \eqref{etat}; we have
			\begin{align}
				\frac{1}{2}\sum_{t=1}^T\eta_t\E_{p_{t,m}}\phi_{t,m}^2=&\frac{1}{2}\sum_{t=1}^{T}\eta_t v_t,\\
				\leq&\frac{\gamma}{2}\sum_{t=1}^{T}\frac{V_t-V_{t-1}}{\sqrt{V_t}},\\
				\leq&\gamma\sum_{t=1}^{T}\sqrt{V_t}-\sqrt{V_{t-1}},\\
				\leq&\gamma\sqrt{V_T},
			\end{align}
		which concludes the proof.
		\end{proof}
	\end{lemma}

	\begin{lemma}\label{thm:ntdt}
		When using \autoref{alg:framework}, $\eta_t$ as in \eqref{etat}, we have
		\begin{align*}
			\sum_{t=1}^{T}\left(1-\frac{\eta_t}{\eta_{t-1}}\right)d_t\leq&\sqrt{V_T+D_T^2}.
		\end{align*}
		\begin{proof}
			From the definitions of $d_t$, $D_t$ and \eqref{etat}; we have
			\begin{align}
				\sum_{t=1}^{T}\left(1-\frac{\eta_t}{\eta_{t-1}}\right)d_t\leq&\sum_{t=1}^{T}\left(1-\frac{\eta_t}{\eta_{t-1}}\right)D_t,\\
				\leq&\sum_{t=1}^{T}\left(1-\frac{\sqrt{V_{t-1}+D_{t-1}^2}}{\sqrt{V_t+D_t^2}}\right)D_t,\\
				\leq&\sum_{t=1}^{T}\sqrt{V_t+D_t^2}-\sqrt{V_{t-1}+D_{t-1}^2},\nonumber\\
				\leq&\sqrt{V_T+D_T^2},
			\end{align}
		which concludes the proof.
		\end{proof}
	\end{lemma}

	\begin{theorem}\label{thm:bound2}
		With the learning rates in \eqref{etat}, we have the following for \autoref{alg:framework}
		\begin{align*}
		\sum_{t=1}^T\E_{p_{t,m}}\phi_{t,m}-\phi_{t,s_t}
		\leq&\frac{W(\Lambda_T)+\gamma}{\gamma}\sqrt{V_T+D_T^2}+{\gamma\sqrt{V_T}},
		\end{align*}
		where $\gamma$ is a user-set parameter and $s_t\triangleq \lambda_{t}(1)$.
		\begin{proof}
			The proof comes from combining \autoref{thm:bound} with \autoref{thm:ntphi2} and \autoref{thm:ntdt} and the definition of $W(\Lambda_T)$.
		\end{proof}
	\end{theorem}
	\autoref{thm:bound2} provides us with a performance bound that is only dependent on a single parameter $\gamma$ which needs to be set at the beginning. However, this does not invalidate the truly online claim since $\gamma$ can be straightforwardly set based on the size of the competition class alone, which is something we naturally have access to at the design of the algorithm. 
	
	\subsection[Designing the Performance Measures]{Designing the Performance Measures $\phi_{t,m}$}
	We set the performance measure $\phi_{t,m}$ as
	\begin{align}
		\phi_{t,m}=&\begin{cases}
			\frac{l_{t,m}-\psi_t}{q_{t,m}}, &m=i_t\\
			0, &m\neq i_t
		\end{cases},\label{phitm}
	\end{align}
	where $l_{t,i_t}$ is the incurred and observed loss at time $t$ from the selection $i_t$ (other losses are hidden because of the bandit setting), $q_{t,m}$ is the selection probability of the arm $m$ at time $t$ and $\psi_t$ is the minimum loss observed so far as
	\begin{align}
		\psi_t=&\min (\psi_{t-1},l_{t,i_t}).\label{psit}
	\end{align}
	To bound the expected regret, we need to bound both of the expectations of $V_t$ and $D_t^2$, which are given by the following results.
	
	\begin{lemma}\label{thm:EVT}
		For $\phi_{t,m}$ as in \eqref{phitm}, $\psi_t$ as in \eqref{psit} and $q_{t,m}$ as in \eqref{qtm}, the expectation of $V_T$ is bounded as follows:
		\begin{align*}
			\E[V_T]&\leq\frac{MT}{\theta}(A-B)^2,
		\end{align*}
		where $[B,A]$ is the range of losses $l_{t,m}$ for all $t,m$ and $\theta\leq 1-\epsilon_t, \forall t$.
		\begin{proof}
			From the definition of $v_t$, we have
			\begin{align}
				v_t&=\frac{p_{t,i_t}(l_{t,i_t}-\psi_t)^2}{q_{t,i_t}^2}.\label{Epvt}
			\end{align}
			From the arm selection probabilities $q_{t,m}$ in \eqref{qtm}, we have
			\begin{align}
				\frac{p_{t,m}}{q_{t,m}}\leq \frac{1}{1-\epsilon_t}\label{p/q}
			\end{align}
			Thus, combining \eqref{Epvt} and \eqref{p/q} gives
			\begin{align}
				v_t&\leq\frac{(l_{t,i_t}-\psi_t)^2}{(1-\epsilon_t)q_{t,i_t}},
			\end{align}
			and its expectation is bounded as
			\begin{align}
				\E[v_t]&\leq\frac{M(A-B)^2}{(1-\epsilon_t)}\leq\frac{M(A-B)^2}{\theta},
			\end{align}
			where $\theta\leq 1-\epsilon_t, \forall t$ and $B\leq l_{t,m}\leq A, \forall t,m$. Thus,
			\begin{align}
				\E[V_T]&=\E[\sum_{t=1}^{T}v_{t}]\leq\frac{MT}{\theta}(A-B)^2,
			\end{align}
			which concludes the proof.
		\end{proof}
	\end{lemma}

	\begin{lemma}\label{thm:EDT}
		For $\phi_{t,m}$ as in \eqref{phitm}, $\psi_t$ as in \eqref{psit} and $q_{t,m}$ as in \eqref{qtm}, the expectation of $D_T^2$ is bounded as follows:
		\begin{align}
			\E[D_T^2]\leq\frac{(A-B)^2}{\epsilon^2}
		\end{align}
		where $[B,A]$ is the range of losses $l_{t,m}$ for all $t,m$ and $\epsilon\leq \epsilon_t/M, \forall t$.
		\begin{proof}
			The proof is straightforward from the definition of $D_T$
			\begin{align}
				D_T&=\max_{1\leq t \leq T}\frac{l_{t,i_t}-\psi_t}{q_{t,i_t}},\\
				&\leq\frac{A-B}{\epsilon},
			\end{align}
			which concludes the proof.
		\end{proof}
	\end{lemma}
	
	\begin{lemma}\label{thm:Epsit}
	For $\phi_{t,m}$ as in \eqref{phitm}, $\psi_t$ as in \eqref{psit} and $q_{t,m}$ as in \eqref{qtm}, we have the following expectation result
	\begin{align*}
		\sum_{t=1}^{T}\E[\psi_t-\min(l_{t,s_t},&\psi_{t-1})]\leq \frac{A-B}{\epsilon},
	\end{align*}
	for any $\{s_t\}_{t=1}^T$ arm selection sequence, where $[B,A]$ is the range of losses $l_{t,m}$ for all $t,m$ and $\epsilon\leq \epsilon_t/M, \forall t$.
	\begin{proof}
	Given $\epsilon\leq {\epsilon_t}/{M}, \forall t$, we have
	\begin{align}
		\E[\psi_t|\psi_{t-1}]\leq \epsilon\min(l_{t,s_t},\psi_{t-1})+(1-\epsilon)\psi_{t-1}
	\end{align}
	Then, the expectation of both sides gives
	\begin{align}
		\E[\psi_t]\leq&\epsilon\E[\min(l_{t,s_t},\psi_{t-1})]+(1-\epsilon)\E[\psi_{t-1}],\\
		\leq&\epsilon\E[\min(l_{t,s_t},\psi_{t-1})]\nonumber\\
		&+(1-\epsilon)\epsilon\E[\min(l_{t-1,m_{t-1}},\psi_{t-2})]\nonumber\\
		&+\ldots\nonumber\\
		&+(1-\epsilon)^{t-2}\epsilon\E[\min(l_{2,m_2},\psi_1)]\nonumber\\
		&+(1-\epsilon)^{t-1}\E[\psi_1].
	\end{align}
	When summed for $t$ from $1$ to $T$, after rearranging, we get
	\begin{align}
		\sum_{t=1}^{T}\E[\psi_t]=&\sum_{t=2}^{T}\epsilon\left(\sum_{\tau=0}^{T-t}(1-\epsilon)^\tau\right)\E[\min(l_{t,s_t},\psi_{t-1})],\nonumber\\
		&+\sum_{t=1}^{T}(1-\epsilon)^{t-1}\E[\psi_1].\label{TEpsit}
	\end{align}
	We define some intermediate variables. Let
	\begin{align}
		Q_t\triangleq\sum_{\tau=0}^{T-t}(1-\epsilon)^\tau,
	\end{align}
	Thus, they have a recursive relation as the following
	\begin{align}
		Q_{t-1}=(1-\epsilon)Q_t+1.
	\end{align}
	Together with \eqref{TEpsit}, we get
	\begin{align}
		\sum_{t=1}^{T}\E[\psi_t]=\sum_{t=2}^{T}\epsilon Q_t\E[\min(l_{t,s_t},\psi_{t-1})]+Q_1\E[\psi_1].
	\end{align}
	Then, we arrive at the intended result by the following
	\begin{align}
		\sum_{t=1}^{T}&\E[\psi_t-\min(l_{t,s_t},\psi_{t-1})]\nonumber\\
		=&\sum_{t=2}^{T}(\epsilon Q_t-1)\E[\min(l_{t,s_t},\psi_{t-1})]+Q_1\E[\psi_1]-l_{1,m_1},\nonumber\\
		=&\sum_{t=2}^{T}(Q_t-Q_{t-1})\E[\min(l_{t,s_t},\psi_{t-1})]+Q_1\E[\psi_1]-l_{1,m_1},\nonumber\\
		\leq&\sum_{t=2}^{T}(Q_t-Q_{t-1})B+Q_1A-B,\\
		\leq&(Q_T-Q_1)B+Q_1A-B,\\
		\leq& Q_1(A-B),\\
		\leq& \frac{A-B}{\epsilon},
	\end{align}
	which concludes the proof.
	\end{proof}
	\end{lemma}
	
	\subsection[Designing the Uniform Mixture Coefficient]{Designing the Uniform Mixture Coefficients $\epsilon_t$}
	We set the uniform mixture coefficients $\epsilon_t$ as the following
	\begin{align}
		\epsilon_t=\min\left(\frac{1}{2},\sqrt{\frac{M}{{t}}}\right).\label{et}
	\end{align}
	From \eqref{et}, the bounds $\theta$ and $\epsilon$ are given by
	\begin{align}
		\frac{1}{\theta}=&2,\\
		\frac{1}{\epsilon}=&\sqrt{MT},
	\end{align}
	when $T\geq 4M$ (for brevity, when $T$ is sufficiently large). Moreover, we also denote the true range with $D$ as 
	\begin{align}
		D\triangleq A-B,
	\end{align}
	where $[B,A]$ is the range of losses $l_{t,m}$ for all $t,m$.
	\begin{lemma}\label{Tet}
		When $\epsilon_{t}$ is as in \eqref{et}, we have the following
		\begin{align*}
			\sum_{t=1}^{T}\epsilon_t\leq 2\sqrt{MT}.
		\end{align*}
		\begin{proof}
			Because of the minimum operation at \eqref{et}, we have
			\begin{align}
				\sum_{t=1}^{T}\epsilon_t\leq&\sum_{t=1}^{T}\sqrt{\frac{M}{t}},\\
				\leq&\sqrt{M}\sum_{t=1}^{T}\frac{(\sqrt{t}-\sqrt{t-1})(\sqrt{t}+\sqrt{t-1})}{\sqrt{t}},\\
				\leq&2\sqrt{MT},
			\end{align}
		which concludes the proof.
		\end{proof}
	\end{lemma}
	
	\begin{corollary}\label{EsqVT}
		For $\epsilon_{t}$ as in \eqref{et}, we have
		\begin{align}
			\E[\sqrt{V_T}]\leq D\sqrt{2MT},
		\end{align}
		where $D$ denotes the true range, i.e., $D\triangleq A-B$.
		\begin{proof}
			The proof comes from the concavity of the square-root, \autoref{thm:EVT} and $\theta=0.5$ (from \eqref{et}).
		\end{proof}
	\end{corollary}
	
	\begin{corollary}\label{EsqDT}
		For $\epsilon_{t}$ as in \eqref{et}, we have
		\begin{align}
			\E[\sqrt{V_T+D_T^2}]\leq D\sqrt{3MT}
		\end{align}
		where $D$ denotes the true range, i.e., $D\triangleq A-B$.
		\begin{proof}
			The proof comes from the concavity of the square-root, \autoref{thm:EVT}, \autoref{thm:EDT} and $\epsilon=1/\sqrt{MT}$ for large $T$ (from \eqref{et}).
		\end{proof}
	\end{corollary}
	
	\begin{corollary}\label{EsqphiT}
		For $\epsilon_{t}$ as in \eqref{et}, we have
		\begin{align}
			\sum_{t=1}^{T}\E[\psi_t-\min(l_{t,s_t},&\psi_{t-1})]\leq D\sqrt{MT},
		\end{align}
		where $D$ denotes the true range, i.e., $D\triangleq A-B$.
		\begin{proof}
			The proof comes from \autoref{thm:Epsit} and $\epsilon=1/\sqrt{MT}$ for large $T$ (from \eqref{et}).
		\end{proof}
	\end{corollary}
	
	\subsection{Expected Regret Bound}
	\begin{theorem}\label{thm:regret}
		For $\eta_t$ as in \eqref{etat}, $\phi_{t,m}$ as in \eqref{phitm}, $\psi_t$ as in \eqref{psit}, $\epsilon_t$ as in \eqref{et} and $q_{t,m}$ as in \eqref{qtm}, we have the following expected regret
		\begin{align*}
			\sum_{t=1}^{T}&\E[l_{t,i_t}]-l_{t,s_t}\leq D\sqrt{MT}\left(3+\sqrt{3}+\frac{W_T}{\gamma}\sqrt{3}+\gamma\sqrt{2}\right),
		\end{align*}
		where $W_T=W(\Lambda_T)$ and $\gamma$ is user-set parameter.
		\begin{proof}
			We have from \eqref{qtm}
			\begin{align}
				\E_{q_{t}}[l_{t,i_t}-\psi_t|\psi_{t-1}]=&(1-\epsilon_t)\E_{p_{t}}[l_{t,i_t}-\psi_t|\psi_{t-1}]\nonumber\\
				&+\epsilon_t\E_u[l_{t,i_t}-\psi_t|\psi_{t-1}],
			\end{align}
			where $\E_u[\cdot]$ is the expectation over uniform distribution. Hence, we have
			\begin{align}
				\E_{q_{t}}[l_{t,i_t}-\psi_t|\psi_{t-1}]\leq&\E_{p_{t}}[l_{t,i_t}-\psi_t|\psi_{t-1}]+\epsilon_t(A-B),\label{qtmptm}
			\end{align}
			where $B\leq l_{t,m}\leq A, \forall t,m$, i.e., $[B,A]$ is the true range of losses.
			Furthermore,
			\begin{align}
				\E&\left[\sum_{t=1}^{T}\E_{p_{t,m}}[\phi_{t,m}]-\phi_{t,s_t}\right]\\
				=&\E\left[\sum_{t=1}^{T}\E_t\left[\E_{p_{t,m}}[\phi_{t,m}]-\phi_{t,s_t}\right]\right],\\
				=&\E\left[\sum_{t=1}^{T}\E_{p_{t}}[l_{t,i_t}-\psi_t|\psi_{t-1}]-l_{t,s_t}+\min(l_{t,s_t},\psi_{t-1})\right],\nonumber\\
				\geq&\E\left[\sum_{t=1}^{T}\E_{q_{t}}[l_{t,i_t}-\psi_t|\psi_{t-1}]-l_{t,s_t}\right]\nonumber\\
				&+\E\left[\sum_{t=1}^{T}\min(l_{t,s_t},\psi_{t-1})-\epsilon_t(A-B)\right],
			\end{align}
			from \eqref{qtmptm}.
			Thus, together with \autoref{thm:bound2}
			\begin{align}
				\sum_{t=1}^{T}\E[l_{t,i_t}]&-l_{t,s_t}\nonumber\\
				\leq& \sum_{t=1}^T \E[\psi_t-\min(l_{t,s_t},\psi_{t-1})]+(A-B)\sum_{t=1}^{T}\epsilon_t\nonumber\\
				&+\E\left[\frac{W(\Lambda_T)+\gamma}{\gamma}\sqrt{V_T+D_T^2}+{\gamma\sqrt{V_T}}\right]
			\end{align}
			Using \autoref{Tet}, \autoref{EsqVT}, \autoref{EsqDT}, \autoref{EsqphiT}; we get
			\begin{align*}
				\sum_{t=1}^{T}\E[l_{t,i_t}]-l_{t,s_t}\leq& D\sqrt{MT}+2D\sqrt{MT}\nonumber\\
				&+\frac{W(\Lambda_T)+\gamma}{\gamma}D\sqrt{3MT}+\gamma D\sqrt{2MT}
			\end{align*}
		Rearranging, we get
		\begin{align*}
			\sum_{t=1}^{T}&\E[l_{t,i_t}]-l_{t,s_t}\leq D\sqrt{MT}\left(3+\sqrt{3}+\frac{W_T}{\gamma}\sqrt{3}+\gamma\sqrt{2}\right),
		\end{align*}
		where $W_T=W(\Lambda_T)$, which concludes the proof.
		\end{proof}
	\end{theorem}
	
	\section{Discussions and Conclusion}\label{sec:conc}
	\begin{corollary}
		When $\gamma=\sqrt{{W}}$, the expected regret against a competition $\{s_t\}_{t=1}^T$ is bounded as
		\begin{align*}
			\sum_{t=1}^{T}&\E[l_{t,i_t}]-l_{t,s_t}\leq D\sqrt{MT}\left(5+4\sqrt{W}\right),
		\end{align*}
		where $W$ is an upper bound on our competing class such that $W(\Lambda_T)\leq W$.
		\begin{proof}
			The proof comes from putting $\gamma=\sqrt{W}$ in \autoref{thm:regret} with some loose bounding.
		\end{proof}
	\end{corollary}

	\begin{remark}
		Hence, we can achieve an expected regret bound of the following order
		\begin{align*}
			\E[R_T] \triangleq& \E\left[\sum_{t=1}^{T}l_{t,i_t}-l_{t,s_t}\right],\\
			=& O\left(D\sqrt{WMT}\right),
		\end{align*}
		where $M$ is the number of bandit arms, i.e., $m\in\{1,\ldots,M\}$; $T$ is the time horizon, i.e., $t\in\{1,\ldots,T\}$; $W$ is an upper bound on our competition complexity, i.e., $W(\Lambda_T)\leq W$; and	$D$ is the unknown range of the losses $l_{t,m}$, i.e.,
		\begin{align*}
			D=\max_{t\in\{1,\ldots,T\}}\max_{m\in\{1,\ldots,M\}}l_{t,m}-\min_{t\in\{1,\ldots,T\}}\min_{m\in\{1,\ldots,M\}}l_{t,m}.
		\end{align*}
	\end{remark}

	\begin{remark}
		Our expected regret bound
		\begin{align*}
			\E[R_T] =O\left(D\sqrt{WMT}\right),
		\end{align*}
		is translation and scale invariant, hence, a fundamental regret bound \cite{cesa2007}.
		\begin{proof}
			Instead of the true loss $l_{t,m}$, let us observe an affine transform $\tilde{l}_{t,m}$ of it, i.e.,
			\begin{align}
				\tilde{l}_{t,m}=\alpha l_{t,m}+\beta,
			\end{align}
			for some $\alpha>0$ and $\beta$. The expected regret bound for these affine transforms would be
			\begin{align}
				\E[\tilde{R}_T]=O\left(\alpha D\sqrt{WMT}\right),
			\end{align}
			because only the range $D$ would be affected. Since this regret is equal to the original regret times $\alpha$, it is translation and scale invariant.
		\end{proof}
	\end{remark}

	In conclusion, we have successfully created a completely online, generalized algorithm for the adversarial multi-armed bandit problem. With suitable design, it is possible to compete against a subset of the all possible bandit arm selection sequences that is appropriate for a given problem. By combining the similar sequences together in each step of the algorithm, and creating appropriate equivalence classes, we can compete against the sequences with minimal redundancy and in a computationally efficient manner. Our performance bounds are translation-free and scale-free of the bandit arm losses.
	
\bibliographystyle{ieeetran}
\bibliography{double_bib}	
	
\end{document}